\documentclass{article} 
\usepackage{iclr2023_conference,times}

\usepackage{amsmath,amsfonts,bm}

\def\eqref#1{equation~\ref{#1}}

\def\1{\bm{1}}

\DeclareMathAlphabet{\mathsfit}{\encodingdefault}{\sfdefault}{m}{sl}
\SetMathAlphabet{\mathsfit}{bold}{\encodingdefault}{\sfdefault}{bx}{n}

\usepackage{hyperref}
\usepackage{url}
\usepackage{cite}
\usepackage{amsmath,amssymb,amsfonts}
\usepackage{graphicx}
\usepackage{textcomp}
\usepackage{xcolor}

\usepackage{booktabs}
\usepackage{bm}
\usepackage{stfloats}
\usepackage{caption}
\usepackage{array}
\usepackage{multirow}
\usepackage{longtable}
\usepackage{rotating}
\usepackage{tabularx}
\newtheorem{definition}{Definition}

\usepackage[linesnumbered,ruled,vlined]{algorithm2e}
\usepackage{algpseudocode} 
\usepackage{bm}
\usepackage{wrapfig}

\title{Graph Contrastive Learning with Personalized Augmentation}

\author{Xin Zhang \thanks{Equal contribution.}\\
Department of Computing\\
The Hong Kong Polytechnic University\\
\texttt{xin12.zhang@connect.polyu.hk} \\
\And
Qiaoyu Tan$^{ *}$ \\
Department of Computer Science \& Engineering \\
Texas A\&M University \\
\texttt{qytan@tamu.edu} \\
\AND
Xiao Huang\\
Department of Computing\\
The Hong Kong Polytechnic University\\
\texttt{xiaohuang@comp.polyu.edu.hk}\\
\And
Bo Li\\
Department of Computing\\
The Hong Kong Polytechnic University\\
\texttt{comp-bo.li@polyu.edu.hk}\\
}

\iclrfinalcopy 
\begin{document}

\maketitle
\begin{abstract}
Graph contrastive learning (GCL) has emerged as an effective tool for learning unsupervised representations of graphs. The key idea is to maximize the agreement between two augmented views of each graph via data augmentation. Existing GCL models mainly focus on applying \textit{identical augmentation strategies} for all graphs within a given scenario. However, real-world graphs are often not monomorphic but abstractions of diverse natures. Even within the same scenario (e.g., macromolecules and online communities), different graphs might need diverse augmentations to perform effective GCL. 
Thus, blindly augmenting all graphs without considering their individual characteristics may undermine the performance of GCL arts.
To deal with this, we propose the first principled framework, termed as \textit{G}raph contrastive learning with \textit{P}ersonalized \textit{A}ugmentation (GPA), to advance conventional GCL by allowing each graph to choose its own suitable augmentation operations.
In essence, GPA infers tailored augmentation strategies for each graph based on its topology and node attributes via a learnable augmentation selector, which is a plug-and-play module and can be effectively trained with downstream GCL models end-to-end. Extensive experiments across 11 benchmark graphs from different types and domains demonstrate the superiority of GPA against state-of-the-art competitors.
Moreover, by visualizing the learned augmentation distributions across different types of datasets, we show that GPA can effectively identify the most suitable augmentations for each graph based on its characteristics.
\end{abstract}

\section{Introduction}
Graph contrastive learning (GCL) could learn effective representations of graphs~\citep{you2021graph} in a self-supervised manner. It attracts considerable attention~\citep{sun2019infograph,peng2020graph,hassani2020contrastive,zhu2021graph,tan2022mgae}, given that labels are not available in many real-world networked systems. The core idea of GCL is to generate two augmented views for each graph by perturbing it, and then learn representations via maximizing the mutual information between the two views~\citep{velickovic2019deep,wan2020contrastive}. Existing efforts can be roughly divided into node-level~\citep{velickovic2019deep,zhu2021graph,wan2020contrastive,peng2020graph} and graph-level~\citep{you2021graph} categories. In this paper, we predominately focus on graph-level contrastive learning research.

The performance of GCL is known to be heavily affected by the chosen augmentation types~\citep{hassani2020contrastive,you2020graph,jin2020self}, since different augmentations may impose different inductive biases about the data. Intensive recent works have been devoted to exploring effective augmentations for different graph scenarios~\citep{velivckovic2018deep,zhu2021graph,you2021graph,chen2020distance}. Typical augmentation strategies include node dropping, edge perturbation, subgraph sampling, and attribute masking. The best augmentation option is often data-driven and varies in graph scales or types~\citep{you2020graph,jin2020self}. For example,~\citep{you2020graph} revealed that edge perturbation may benefit social networks but hurt biochemical molecules. Therefore, manually searching augmentation strategies for a given scenario would involve extensive trials and efforts, hindering the practical usages of GCL. Several studies have been proposed to address this issue by automating GCL.

Despite the recent advances, existing GCL might be suboptimal for augmentation configuration, since they apply a well-chosen but identical augmentation option to all graphs in a dataset~\citep{you2020graph,you2021graph}. The rationale is: graphs in a scenario usually have different properties because the characteristics of real-world graphs are complex and diverse~\citep{west2001introduction,liu2019single}. For example, by slightly changing the structure of a molecular graph, its target function could be completely different~\citep{Morris+2020}. Similar observations have also been found in many other graph domains, such as social communities and protein-protein interaction networks~\citep{thakoor2021bootstrapped}. 
Meanwhile, it has been demonstrated that, even when a single augmentation method (e.g., edge dropping) is used, we could still learn personalized dropping strategies for different graphs to improve the performance of GCL~\citep{suresh2021adversarial}. Motivated by these observations, a natural question arises: \textit{What are the impacts of different augmentation strategies on a given graph (an instance in a graph dataset)? Can we build a stronger automated GCL by allowing each graph instance to choose its favorable augmentation types?}

\begin{figure}[t!]
    \centering
  \includegraphics[width=10cm]{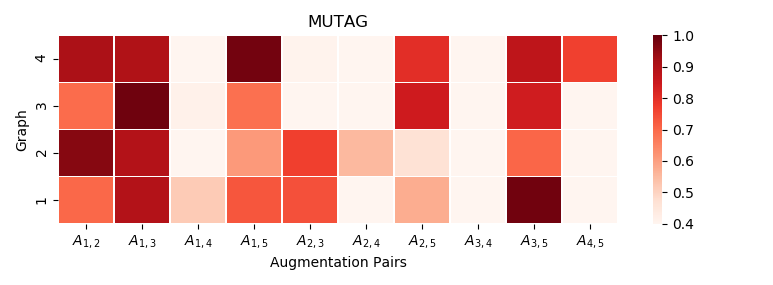}
  \vspace{-0pt}
  \caption{The effect of different augmentation strategies toward four randomly sampled graphs from MUTAG. The results are obtained by training GraphCL ten times with different augmentation choices. X-axis denotes the id of augmentation pair. Y-axis is the graph id. 
  The color represents the performance. 
  The darker the color is, the better performance GCL achieves under the corresponding augmentation strategy.
  }
  \label{figure_example_intro}
  \vspace{-0cm}
\end{figure}

To understand the first question, we conduct a preliminary experiment on the MUTAG dataset to test how different augmentation types impact the GCL results in Fig.~\ref{figure_example_intro}. In general, we found that different graphs favor distinct perturbation operations to achieve their best performances. For example, the first graph performs better when using \textit{edge perturbation \& attribute masking} as augmentations, while the second graph prefers \textit{identical \& node dropping} combination. Such personalized phenomenon of graph instances, in terms of their desirable augmentation strategies, has never been explored in GCL. To bridge the gap, in this paper, we propose to develop an effective augmentation selector to identify the most informative perturbation operators for different graphs when performing GCL on a specific dataset.

However, it is a challenging task to perform personalized augmentations in GCL due to two major roadblocks. First, unlike the traditional GCL setting, the search space in our personalized scenario grows exponentially w.r.t. the number of graphs in a dataset. For instance, let $K$ denote the number of augmentation types of interest and $N$ be the number of graphs, the candidate space for each graph is ${K+1 \choose 2}$, 
leading to ${K+1 \choose 2}^N$ combination options for all graphs in a dataset. This search space is intolerable in practice since $N$ could be thousands or even tens of thousands~\citep{hu2020open}. As a result, the conventional trial-and-error approach cannot be directly applied because each trail itself (i.e., testing an augmentation option for a dataset) is time-consuming. Second, the augmentation choices and GCL model naturally depend on and reinforce each other since the contrastive loss consists of augmented views and GCL encoder. That is, learning a better GCL requires well-chosen augmentation strategies~\citep{you2020graph}, while selecting suitable augmentation operators needs the signals of GCL as feedback~\citep{you2021graph}. Thus, how to perform effective personalized augmentation selector on a premise of such mutual effect is another challenge.

To tackle the aforementioned challenges, we propose a novel contrastive learning framework, dubbed GPA. Specifically, GPA works by iteratively updating a personalized augmentation selector and the GCL method, where the former aims to identify optimal augmentation types for each graph instance, and the latter is trained according to an instance-level contrastive loss defined on those assigned augmentation options. To further illustrate the difference between the existing GCL methods and our model GPA, we outline the properties of them in Table~\ref{GCL_overview}, where GPA dominates in the comparisons.
We summarize our main contributions as follows.

\begin{itemize}
    \item We focus on augmentation selection for graph contrastive learning, and propose an effective personalized augmentation framework (GPA). 
    To the best of our knowledge, GPA is the first to assign personalized augmentation types to each graph based on its own characteristic when performing GCL. Moreover, we automate the personalized augmentation process, equipping GPA with broader applicability and practicability.
    \item To cope with the huge search space, we develop a personalized augmentation selector to effectively infer optimal augmentation strategies by relaxing the discrete space to be continuous. 
    To exploit the mutual reinforced effect between augmentation strategies and GCL, GPA is formulated as a bi-level optimization, which learns the augmentation selector and GCL method systematically.
    \item We conduct extensive experiments to evaluate GPA on multiple graph benchmarks of diverse scales and types. Empirical results demonstrate the superiority of GPA against state-of-the-art GCL competitors. 
\end{itemize}

\begin{table}[t]
\centering
  \caption{An overview of graph contrastive learning methods. Our GPA model is the first automated GCL effort that supports personalized augmentation configuration in terms of various augmentation types. 
  }
  \label{GCL_overview}
    \vspace{-0pt}
     \begin{small}
\setlength{\tabcolsep}{2.0pt}
  {
  \begin{tabular}{lcccccc}
   \toprule
     &Property &GraphCL
     &JOAO
     &AD-GCL
     &Ours\\
     \hline
     &NodeDrop
     &$\checkmark$
     &$\checkmark$
     &-
     &$\checkmark$
     \\
     &EdgePert
     &$\checkmark$
     &$\checkmark$
     &$\checkmark$
     &$\checkmark$
     \\
     &AttMask
     &$\checkmark$
     &$\checkmark$
     &-
     &$\checkmark$
     \\
     &Subgraph
     &$\checkmark$
     &$\checkmark$
     &-
     &$\checkmark$
     \\
     &Auto
     &-
     &$\checkmark$
     &$\checkmark$
     &$\checkmark$
     \\
     &Personalized
     &-
     &-
     &$\checkmark$
     &$\checkmark$
     \\
 \bottomrule
\end{tabular}}
\end{small}
\vspace{0pt}
\end{table}

\section{Related Work}
\label{related_work}
There are two lines of research that are closely related to our work, namely, graph representation learning and graph contrastive learning. In this section, we briefly
review some related works in these two fields. Please refer to~\citep{liu2021self} and~\citep{wu2020comprehensive} for comprehensive review. 

\noindent
\textbf{Graph representation learning.} With the rapid development of graph neural networks (GNNs), a large number of GNN-based graph representation learning frameworks have been proposed~\citep{wu2020comprehensive,wang2019learning,iyer2021bi,tan2020learning,wang2022augmentation,tan2019deep}, which exhibits promising performance. Typically, these methods can be divided into supervised and unsupervised categories. While supervised methods~\citep{chen2018fastgcn,ding2018semi,kipf2017semi,velivckovic2017graph} achieve empirical success with the help of labels, reliable labels are often scarce in real-world scenarios. Thus, unsupervised graph learning approaches~\citep{kipf2016variational,garcia2017learning,hamilton2017inductive,hou2022graphmae} have broader application potential. For example, one of the well-known methods is GAE~\citep{kipf2016variational} which learns graph representations by reconstructing the network structure under the autoencoder approach. Another popular approach GraphSAGE~\citep{hamilton2017inductive} aims to train GNNs by a random-walk based objective.

\noindent
\textbf{Graph contrastive learning.}
Graph contrastive learning (GCL) has attracted significant attention in the past two years for self-supervised graph learning~\citep{wu2021self}.
It learns GNN encoder by maximizing the agreement between representations of a graph in its different augmented views, so that similar graphs are close to each other, while dissimilar ones are spaced apart. Many GCL efforts have been devoted to  
node level~\citep{wan2020contrastive,peng2020graph,hu2020gpt}, subgraph level~\citep{qiu2020gcc,jiao2020sub}, and graph level~\citep{you2020graph,zeng2020contrastive} scenarios, where the key challenge lies in designing effective graph augmentations. Recently, GraphCL~\citep{you2020graph} introduced four types of graph augmentations, including node dropping, edge perturbation, sub-graph sampling, and node attribute masking, and showed that different graph applications may favor different augmentation combinations. However, the optimal augmentation configuration for a given dataset is mainly determined by either domain experts or extensive trial-and-errors, thus limiting the boarder applications of GCL in practice. 

More recently, JOAO~\citep{you2021graph} proposes to directly learn the sampling distribution of four graph augmentation types for a given dataset from the data itself. Although the sampled augmentation types for different graphs may be different to some extent, due to the randomness of sampling, their deviations might be trivial as the distribution vector is shared for all sample graphs. In view of this, JOAO is not a personalized approach because its goal is not to learn tailored augmentation types for different graphs. AD-GCL~\citep{suresh2021adversarial} develops a learnable edge-dropping augmentation via adversarial training. It can generate different edge-dropping structures for different graphs. Nevertheless, AD-GCL only supports edge-dropping graph augmentation, and cannot be trivially adopted to other graph augmentation types, such as node attribute masking.

In this work, we build a learnable augmentation type selector that identifies the most informative augmentation types for each sample graph. Compared to the existing GCL methods, our method is more general because it supports arbitrary augmentation types. Moreover, it is end-to-end differentiable and can be jointly trained with downstream GCL algorithms. An overview of the existing literature of graph contrastive learning methods is summarized in Table~\ref{GCL_overview}.

\section{Notations and Preliminaries}
\label{preliminaries}
We first introduce the notations used throughout this paper and preliminary concepts for further exposition. Let $\mathcal{G}=\{\mathcal{G}_n: 1\leq n \leq N\}$ denote a graph dataset with $N$ sample graphs, where $\mathcal{G}_n=(\mathcal{V},\mathcal{E})\in\mathcal{G}$ stands for an undirected graph with nodes $\mathcal{V}$ and edges $\mathcal{E}$. Each node $v\in\mathcal{V}$ in $\mathcal{G}_n$ is described by an $F$-dimensional feature vector $\mathbf{X}_v\in\mathbb{R}^F$. We use $\mathcal{A}=\{A_k: 1\leq k \leq K\}$ to denote a set of data augmentation operators, where $K$ is the maximum number of augmentation types of interest. Each augmentation operator $A_k: \mathcal{G}_n \rightarrow{} \Tilde{\mathcal{G}}_n$ transforms a graph into its conceptually similar form with certain prior. In previous GCL studies, they focus on identifying two optimal augmentation types for the whole dataset $\mathcal{G}$, denoted by augmentation pair $A_{i,j}=(A_i,A_j)$, where $A_i,A_j\in\mathcal{A}$. The optimal augmentation pair here is often manually picked via rules of thumb or trial-and-error. However, as shown in Fig.~\ref{figure_example_intro}, different graphs within the dataset may favor different augmentation pairs. Therefore, we propose personalized augmentation selection to assign suitable augmentation pairs for different graphs, where the problem is defined below.

\begin{definition}
\textbf{Personalized augmentation selection.} Given a graph dataset $\mathcal{G}=(\mathcal{G}_n: 1\leq n \leq N)$, and the augmentation space $\mathcal{A}=\{A_k: 1\leq k \leq K\}$ consisting of $K$ different augmentation types, to perform GCL, personalized augmentation selection aims to find the optimal augmentation pair $A_{i,j}^n=(A_i^n, A_j^n)$ for each graph $\mathcal{G}_n\in\mathcal{G}$. The values of $i$ and $j$ are only determined by the characteristic of the $n$-th sample graph $\mathcal{G}_n$.
\label{def_pas}
\end{definition}

\begin{figure*}[h]
  \centering
  \includegraphics[width=13cm]{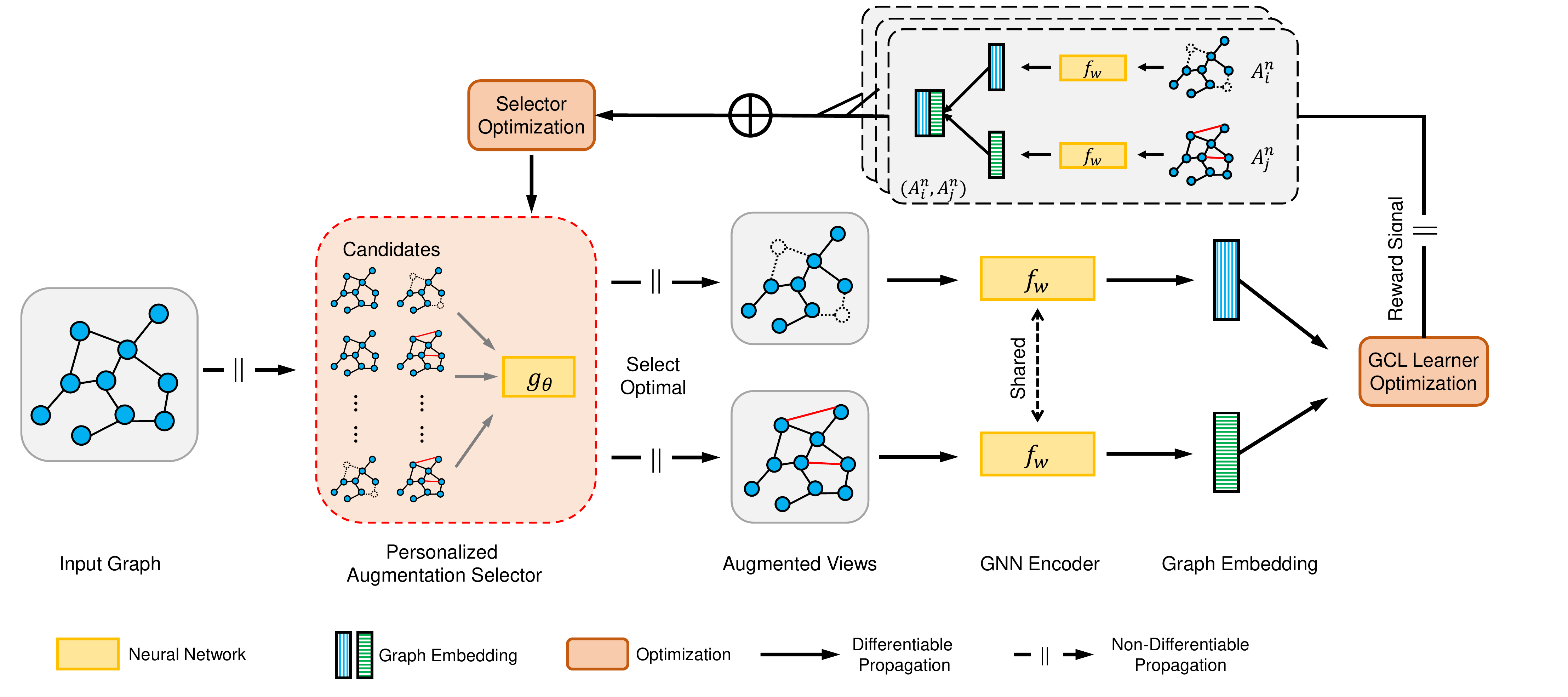}
  \vspace{-0pt}
  \caption{Illustration of our GPA framework. The \textit{personalized augmentation selector} infers the two most informative augmentation operators, and the \textit{GCL model} trains GNN encoder based on the sampled augmented views. Specifically, the \textit{personalized augmentation selector} is learned to adjust its selection strategy to infer optimal augmentations on each graph, according to the characteristic of each graph and the \textit{GCL model}'s performance, i.e., loss.}
  \label{figure_flowchart}
\end{figure*}

\subsection{Graph Representation Learning}
Given a graph $\mathcal{G}_n=(\mathcal{V},\mathcal{E})$, the goal of graph representation learning is to learn a mapping function $f_w: \mathcal{G}_n\xrightarrow{} \mathbb{R}^D$ to encode each sample graph $\mathcal{G}_n\in\mathcal{G}$ into a $D$-dimensional vector. Inspired by the impressive ability of graph neural networks (GNNs) in modeling graph-structured data, GNNs have become the default tools to implement the mapping function. Following the message passing strategy~\citep{gilmer2017neural}, the core idea of GNNs is to update the representation of each node by capturing the structure information within its neighborhood. Considering a $K$ layer GNN $f_w(\cdot)$, the $k$-th layer propagation rule is represented as:
\begin{equation}
    \mathbf{h}_v^{(k)}=\text{UPDATE}^{(k)}\left(\mathbf{h}_v^{(k-1)}, \text{AGG}^{(k)}(\{\mathbf{h}_u^{(k-1)}:u\in \mathcal{N}_v\})\right),
\label{GNN_propagation}
\end{equation}
where $\mathbf{h}_v^{(k)}\in\mathbb{R}^D$ is the embedding of node $v$ with $\mathbf{h}_v^{(0)}=\mathbf{X}_v$, and $\mathcal{N}_v=\{u: (v,u)\in\mathcal{E}\}$ denotes the neighbor set of node $v$. The $\text{AGG}$ function aims to aggregate messages from the neighbors, while the $\text{UPDATE}$ function targets to update $v$'s embedding based on the aggregated information and its embedding from previous layer. After the $K$ layer propagation, the final embedding for $\mathcal{G}_n$ is summarized over all node embeddings through the READOUT function, expressed as:
\begin{equation}
    \mathbf{h}_n=\text{READOUT}(\{\mathbf{h}_v^{(K)}: v\in\mathcal{G}_n\}),
\end{equation}
where $\mathbf{h}_n\in\mathbb{R}^D$ is the embedding vector of the graph $\mathcal{G}_n$.

\section{Methodology}
In this section, we present the details of the proposed GPA shown in Fig.~\ref{figure_flowchart}.
In a nutshell, it contains two critical components: the \textit{personalized augmentation selector} and the \textit{GCL model}. The former module aims to infer augmentation choices for the downstream GCL methods when training them on the training set, while the later provides reward to update the augmentation selector based on the validation set.  
In the following, we first illustrate the exponential selection space of our personalized augmentation setting. Then, we elaborate the details of the augmentation selector and the GCL method. Finally, we show how to jointly optimize the two components in a unified perspective.

\subsection{Personalized Augmentation Selection Space}
\label{search_space}
Given the graph dataset $\mathcal{G}=\{\mathcal{G}_n: 1\leq n \leq N\}$ and a pool of augmentation operators $\mathcal{A}=\{A_1,A_2,\cdots,A_K\}$, existing GCL efforts aim to select two informative operators (e.g., $(A_i, A_{j}\mid1\leq i,j\leq K)$) for $N$ graphs to create their augmented views. Since the augmentation operators are shared for the whole dataset, the total selection space is ${K+1 \choose 2}$, i.e., unordered sampling two operators with replacement. This collective selection strategy is widely adopted in existing GCL works. However, as discussed before, various sample graphs may favor different augmentation operators owing to the diversity of graph-structured data. Therefore, we propose to adaptively choose two augmentation strategies for different graphs. Following Definition~\ref{def_pas}, we define the selected augmentations for each graph $\mathcal{G}_n$ as $A^n_{i,j}=(A_i^n, A_j^n)$. Then, the potential augmentation selection size for each graph is ${K+1 \choose 2}$, and the total selection space for the whole dataset equals to ${K+1 \choose 2}^N$. Although $K$ is empirically small (e.g., $K=5$) in GCL domain, the total selection space in our personalized setting is still huge and intractable, since the complexity grows exponentially to the number of graphs. For instance, when $K=5$ and $N=100$, we already have $15^{100}$ selection configurations roughly. The situation is more serious in real-world scenarios where $N$ is thousands or even tens of thousands. In this paper, we adopt five essential augmentation operators (i.e. $K=5$), denoted by $\mathcal{A}=\{\text{Identical}, \text{NodeDrop}, \text{EdgePert},\text{Subgraph},\text{AttMask}\}$. These augmentations are initially proposed by the pioneering work~\citep{you2020graph}, and have been demonstrated to be effective for contrastive learning~\citep{you2021graph}. The augmentation pairs are shown in Table~\ref{aug_pairs}, and the details of these augmentations are listed below.

\begin{itemize}
    \item \textbf{NodeDrop.} Given the graph $\mathcal{G}_n$, NodeDrop randomly discards a fraction of the vertices and their connections. The dropping probability of each node follows the i.i.d. uniform distribution. The underlying assumption is that missing part of vertices does not damage the semantic information of $\mathcal{G}_n$.
    \item \textbf{EdgePert.} The connectivity in $\mathcal{G}_n$ is perturbed through randomly adding or dropping a certain portion of edges. We also follow the i.i.d. uniform distribution to add/drop each edge. The underlying prior is that the semantic meaning of $\mathcal{G}_n$ is robust to the variance of edges.
    \item \textbf{AttMask.} AttMask masks the attributes of a certain proportion of vertices. Similarly, each node's masking possibility follows the i.i.d. uniform distribution. Attribute masking implies that the absence of some vertex attributes does not affect the semantics of $\mathcal{G}_n$.
    \item \textbf{Subgraph.} This augmentation method samples a subgraph from the given graph $\mathcal{G}_n$ based on random walk. It believes that most of the semantic meaning of $\mathcal{G}_n$ can be preserved in its local structure.
\end{itemize}

In summary, by considering personalized augmentation, the search space per dataset increases from $ {K+1 \choose 2}$ to $ {K+1 \choose 2}^N$. Therefore, common selection techniques such as rules of thumb or trial-and-errors adopted by prior GCL approaches~\citep{you2021graph,you2020graph} are no longer appropriate. Thus, a tailored augmentation selector is needed to effectively tackle the challenging personalized augmentation problem.

\subsection{Personalized Augmentation Selector}
\label{inductive_search}
In order to assign different augmentation operators to various sample graphs when performing GCL on a specific dataset, random selection is the intuitive solution. Its key idea is to randomly sample two augmentation types for each graph from the candidate set. Despite the simplicity, the random selection approach fails to control the quality of sampled augmentation operators. Therefore, directly coupling the existing GCL framework with random augmentation selection would lead to performance degradation (shown as the variant: GPA-random in Sec.~\ref{ablation}).

To address this issue, we focus on data-driven search by making the augmentation selection process learnable. The principle idea is to parameterize our personalized augmentation selector with a deep neural network, which takes a query graph as input and outputs its optimal augmentation choices. There are two main hurdles to achieve this goal: (i) given the exponential augmentation space, how can we make our personalized augmentation selector scale to the real-world dataset with thousands of graphs; 
(ii) since the topology structure and node attributes are crucial to graph-structured data, how can our personalized augmentation selector exploit this information to produce a more precise augmentation choice?
We illustrate our dedicated solutions below.

Given the augmentation pool $\mathcal{A}=\{A_i: 1\leq i \leq K\}$ and a query graph $\mathcal{G}_n$, our augmentation selector is required to select the most two informative augmentations, e.g., $(A_1^n, A_3^n)$, from the candidate set. This selection problem is well-known to be discrete and non-differentiable. Although enormous efforts based on evolution or reinforcement learning have been proposed to address the discrete selection problem, they are still not suitable for such a large selection space (illustrated in Sec.~\ref{search_space}) and are far from utilizing the properties of graphs. Therefore, we propose to make the search space learnable by relaxing the discrete selection space to be continuous inspired by~\citep{liu2018darts} and further make this relaxation consider the characteristic of graphs. Specifically, given the sampled augmentation pair $A^n_{i,j}=(A^n_i,A^n_j)$ of $\mathcal{G}_n$, its importance score $\hat{\alpha}^n_{i,j}$ is computed as
\begin{equation}
\begin{aligned}
     \hat{\alpha}^n_{i,j}&=\frac{\exp(\alpha_{i,j}^n)}{\sum_{i^\prime j^\prime}\exp(\alpha^n_{i^\prime, j^\prime})},\\
     \alpha_{i,j}^n&=g_\theta(f_w(A^n_i(\mathcal{G}_n)\parallel A^n_j(\mathcal{G}_n))),
\end{aligned}
\label{eq:continuous}
\end{equation}
where $g_\theta$ denotes the score function that takes the representations of augmented views $A^n_i(\mathcal{G}_n)$ and $A^n_j(\mathcal{G}_n)$ as input. This design enforces the score estimation to take into account the topology and node attributes of $\mathcal{G}_n$. In practice, $g_\theta$ is parameterized as a two-layer MLP with a ReLU activation function. $f_w$ is the GNN encoder for graph representation learning. $\parallel$ indicates the concatenation operation. Through \eqref{eq:continuous}, the discrete augmentation selection process reduces to learning a score function $g_{\theta}$ under the consideration of graph characteristic.

After the personalized selector is well-trained, let $\alpha^n=[\hat{{\alpha}}^n_{1,1},\cdots,\hat{\alpha}^n_{i,j},\cdots,\hat{\alpha}^n_{K,K}]$ denote the vector of importance scores associated with all different augmentation pairs of the graph $\mathcal{G}_n$. The optimal augmentation choice of $\mathcal{G}_n$ can be obtained by selecting the augmentation pair with the maximum score in $\alpha^n$. For example, $A^n=(A^n_i, A^n_j)$ if $\hat{\alpha}^n_{i,j}=\arg\max_{i^\prime, j^\prime} \hat{\alpha}^n_{i^\prime, j^\prime}$.

To summarize, the above equation provides a principled solution to our personalized augmentation setting. On one hand, it allows augmentation selection in such a large search space via the simple forward propagation of a shallow neural network, i.e., $g_\theta$. On the other hand, it can also infer the most informative augmentations for each graph based on its own characteristic for downstream GCL model training(discussed in Sec.~\ref{sec-gcl}).

\subsection{GCL Model Learning}
\label{sec-gcl}
After the personalized augmentation selector is well-learned, we can adopt it to train the GCL model. Notice that our model is applicable to arbitrary GCL methods that rely on two augmented views as input. In this section, we mainly focus on the most popular and generic GCL architecture - GraphCL~\citep{you2020graph} as the backbone and leave other specific architectures for future work.

Assume $(A_i^n, A^n_j)$ is the optimal augmentation pair of graph $\mathcal{G}_n$ identified by our personalized augmentation selector, and $f_w(\cdot)$ is the GNN encoder. GraphCL proposes to learn $f_w(\cdot)$ by maximizing the agreement between the two augmented views, i.e., $A_i^n(\mathcal{G}_n), A^n_j(\mathcal{G}_n)$. As discussed in Sec.~\ref{related_work}, the GNN encoder can encode the whole graph into a hidden space $\mathbb{R}^D$. In practice, a shared projection head function $\mathbb{R}^D\rightarrow \mathbb{R}^D$ is often applied upon the output of GNN encoder to improve the model capacity. In the following sections, we abuse the notation $f_w$ to denote both the GNN encoding function and the projection function. Based on this notation, we can formally calculate the instance-level contrastive loss as follows:

\begin{equation}
\small
    \mathcal{L}(\mathcal{G}_n)=-\log\frac{\exp(\text{sim}(f_w(A^n_i(\mathcal{G}_n)), f_w(A^n_j(\mathcal{G}_n)))/\tau)}{\sum_{n^\prime=1,n^\prime\neq n}^{N}\exp(\text{sim}(f_w(A^n_i(\mathcal{G}_n)), f_w(A^{n^\prime}_j(\mathcal{G}_{n^\prime})))/\tau)},
\label{loss_n}
\end{equation}
where $\text{sim}(\cdot,\cdot)$ denotes the cosine similarity function and $\tau$ is the temperature parameter. By minimizing \eqref{loss_n}, it encourages the two augmented views of the same sample graph to have similar representations, while enforces the augmented representations of disparate graphs to be highly distinct. As the sum operation over all graphs $\mathcal{G}$ in the denominator of \eqref{loss_n} is computationally prohibitive, GCL is often trained under minibatch sampling~\citep{you2020graph}, where the negative views are generated from the augmented graphs within the same minibatch.

Although \eqref{loss_n} looks similar to the traditional contrastive loss, the key difference between them is that the augmentation operators $A_i^n, A_j^n$ are closely related to the sample graph $\mathcal{G}_n$ in our formulation. That is, the augmentation operators learned by our model vary from one graph to another according to their own characteristics, which are previously enforced to be the same regardless of the graph's diverse nature. Benefiting from considering the graph's personality, the GCL method can learn the basic but essential features of graphs and thus achieve more expressive representations for downstream tasks, as empirically verified in Sec.~\ref{experiment}.

\subsection{Model Optimization}
\label{optimization}
Until now, we have illustrated the detailed personalized augmentation selector as well as the downstream GCL framework,
the remaining question is how to effectively train the two modules. The naive solution is to first train the personalized augmentation selector separately, and then optimize the GCL method using the identified personalized augmentations as input.
However, such a task-agnostic approach is suboptimal, since it does not take the mutual reinforced effect between augmentation and the GCL method into consideration. This is because learning a better GCL method requires optimal augmentation strategies since they can augment more personalized features to distinguish it from other objects. Meanwhile, obtaining suitable augmentation strategies also needs the signals of a better GCL method as guidance for optimization.
As a result, without linking the two modules in a principled way, it is almost infeasible to enforce the \textit{personalized augmentation selector} to accurately infer optimal augmentation strategy for improving the performance of the \textit{GCL method}. To this end, we propose to tackle this problem by jointly training the two modules under a bi-level optimization, expressed as:

\begin{equation}
\begin{split}
&\arg\min\limits_{\theta} \,\, 
\mathcal{L}_{valid}(w^*(\theta),\theta)\\
&s.t.\quad w^*(\theta)=\mathop{\arg\min}\limits_{w}\mathcal{L}_{train}(w,\theta),
\end{split}
\label{eq:bilevel}
\end{equation}
where $\theta$ denotes the trainable parameters of the personalized augmentation selector, and $w$ is the parameters for the GCL method. The upper-level objective $\mathcal{L}_{valid}(w^*(\theta), \theta)$ aims to find $\theta$ that minimizes the validation rewards on the validation set given the optimal $w^*$, and the lower-level objective $\mathcal{L}_{train}(w, \theta^*)$ targets to optimize $w$ by minimizing the contrastive loss based on the training set with $\theta$ fixed.
We want to remark that GPA only exploits the signals from the self-supervisory task itself without accessing labels. Thus, compared with conventional supervised methods, the validation set here only contains a set of graphs without label information, which is much easier to construct, e.g., randomly sampling 10\% of the training set.

By optimizing \eqref{eq:bilevel}, the personalized augmentation selector and the target GCL model will be jointly trained to reinforce their reciprocal effects. Since deriving exact solutions for this bi-level problem is indeed analytically intractable, we adopt the alternating gradient descent algorithm to solve it as follows.

\subsubsection{Lower-level optimization.} With $\theta$ fixed, we can update $w$ with the standard gradient descent procedure as bellow.
\begin{equation}
    w^{\prime}=w-\xi \nabla_{w}\mathcal{L}_{train}(w,\theta),
\label{encoder_update}
\end{equation}
where $\mathcal{L}_{train}=\mathbb{E}_{\mathcal{G}_{train}} \mathcal{L}(\mathcal{G}_n)$ with $\mathcal{L}(\mathcal{G}_n)$ is instance-level contrastive loss defined in \eqref{loss_n}, $\mathcal{G}_{train}$ denotes the training set, and $\xi$ is the learning rate.

\subsubsection{Upper-level optimization.} 
Since it is not intuitive to directly calculate the gradient \textit{w.r.t.} $\theta$ over all augmentation options, we first define the upper-level objective based on \eqref{loss_n} as below:

\begin{equation}
\begin{split}
\mathcal{L}_{valid}(w^*(\theta),\theta)=\sum_{A_i,A_j\in\mathcal{A}} \sum_{\mathcal{G}_n\in\mathcal{G}_{valid}} \hat{\alpha}^n_{i,j}\mathcal{L}(\mathcal{G}_n),
\end{split}
\label{loss_meta}
\end{equation}
where $\hat{\alpha}^n_{i,j}$ is the selecting score computed by the score function $g_\theta$ with parameter $\theta$ defined in Eq.~\eqref{eq:continuous}. $\mathcal{G}_{valid}$ denotes the validation set. Based on the above loss function, we can update $\theta$ by fixing $w$, expressed as:
\begin{equation}
    \theta^{\prime}=\theta-\xi\nabla_{\theta}\mathcal{L}_{valid}(w^*(\theta),\theta).
\label{weights_update1}
\end{equation}

However, evaluating the gradient \textit{w.r.t.} $\theta$ exactly is intractable and computationally expensive, since it requires solving for the optimal $w^*(\theta)$ whenever $\theta$ gets updated. To approximate the optimal solution $w^*(\theta)$, we propose to take one step of gradient descent update for $w$, without solving the lower-level optimization completely by training until convergence. To further compute the gradient of $\theta$, we apply chain rule to differentiate $\mathcal{L}_{train}(w^\prime(\theta), \theta)$ with respect to $\theta$ via $w^\prime$, where $w^\prime$ is defined in \eqref{encoder_update}.

Therefore, the gradient of $\theta$ can be approximated as:
\begin{equation}
\begin{aligned}
        \nabla_{\theta}\mathcal{L}_{valid}(w^*(\theta),\alpha)
        &\approx\nabla_{\theta}\mathcal{L}_{valid}(w^{\prime},\theta)\\
        &\approx \nabla_{\theta}\mathcal{L}_{valid}(w-\xi \nabla_{w}\mathcal{L}_{train}(w,\theta),\theta).\\
\end{aligned}
\label{simple_approx}
\end{equation}
Applying chain rule to \eqref{simple_approx}, we further obtain the gradient of $\theta$ as:
\begin{equation}
\begin{aligned}
        \nabla_{\theta}\mathcal{L}_{valid}(w^*(\theta)\theta)
        &\approx \nabla_{\theta}\mathcal{L}_{valid}(w^{\prime},\theta)\\&-\xi\nabla^{2}_{\theta,w}\mathcal{L}_{train}(w,\theta)\nabla_{w^{\prime}}\mathcal{L}_{valid}(w^{\prime},\theta)\\
\end{aligned}
\label{simple_approx2}
\end{equation}

Since the computation cost of the second term in \eqref{simple_approx2} is still high, it can be further approximated by the finite difference method:

\begin{equation}
\begin{aligned}
    \xi\nabla^{2}_{\theta,w}&\mathcal{L}_{train}(w,\theta)\nabla_{w^{\prime}}\mathcal{L}_{valid}(w^{\prime},\theta)\\
    &\approx\frac{\nabla_{\theta}\mathcal{L}_{train}(w^+,\theta)-\nabla_{\theta}\mathcal{L}_{train}(w^-,\theta)}{2\epsilon},
\end{aligned}
\label{simple_approx3}
\end{equation}
where $w^{\pm}=w\pm\epsilon \nabla_{w^{\prime}}\mathcal{L}_{valid}(w^{\prime},\theta)$ and $\epsilon$ denotes a small scalar.

The final result of the gradient of $\theta$ is:
\begin{equation}
    \begin{split}
        \nabla_\theta \mathcal{L}_{valid}(w^*(\theta), \theta)&\approx \nabla_\theta \mathcal{L}_{valid}(w^\prime, \theta) \\
        &- \xi\frac{\nabla_{\theta}\mathcal{L}_{train}(w^+,\theta)-\nabla_{\theta}\mathcal{L}_{train}(w^-,\theta)}{2\epsilon}.
    \end{split}
\label{weights_update2}
\end{equation}

\begin{figure}[h]
    \centering
  \includegraphics[width=8.5cm]{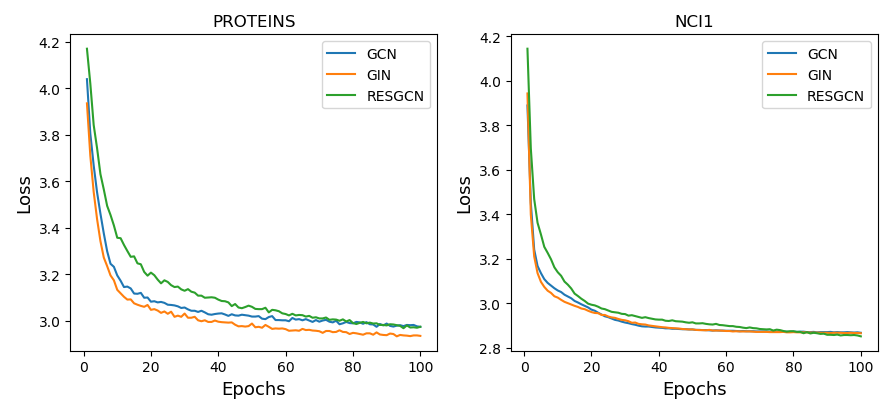}
  \vspace{-6pt}

  \caption{Empirical training curves of approximate gradient scheme in GPA on datasets PROTEINS and NCI1 with different GNN encoders.}
  \label{figure_loss_curve}
  \vspace{-6pt}
\end{figure}

By alternating the update rules in \eqref{encoder_update} and \eqref{weights_update1}, we are able to progressively learn the two modules. Although an optimizer with the theoretical guarantee of convergence for the bi-level problem in \eqref{eq:bilevel} remains an open challenge, alternating gradient descent algorithm has been widely adopted to solve similar objectives in Bayesian optimization~\citep{snoek2012practical}, automatic differentiation~\citep{shaban2019truncated}, and adversarial training~\citep{wang2019towards}. The complete optimization procedure of our model is shown in Algorithm~\ref{alg1}. We also show some level of empirical convergence as seen in Fig.~\ref{figure_loss_curve}.

\begin{algorithm}[tb]
\DontPrintSemicolon
  \caption{The framework of GPA}
\label{alg1}
    \KwIn{A graph dataset $\mathcal{G}$, the personalized augmentation selector $g_{\theta}(\cdot)$, and a GCL model $f_w(\cdot)$;}
    \KwOut{The well-trained GCL model;}
    Split the input graph dataset $\mathcal{G}$ into train $\mathcal{G}_{train}$ and validation $\mathcal{G}_{valid}$ set;
    
    Initialize the selector parameter $\theta$ and the GCL model parameter $w$;
    
    \While{not converge}{
    Randomly sample a minibatch of graphs from the training set;
    
    Infer the optimal augmentation pairs for sampled graphs using the \textit{personalized augmentation selector} $g_\theta(\cdot)$;
    
    Update parameters $w$ of the \textit{GCL learner} based on the sampled graphs and the identified augmentation types according to Eq.~\eqref{encoder_update};
    
    Randomly sample a batch of graphs from validation set;
    
    Compute the rewards based on the sampled validation graphs using the updated $w^\prime$ according to Eq.~\eqref{loss_meta};
    
    Update parameters $\theta$ of the \textit{personalized augmentation selector} according to Eq.~\eqref{weights_update1} and Eq.~\eqref{weights_update2};
    }
    {\bf Return} The well-trained \textit{GCL model}.
\end{algorithm}

\begin{table}[t]
\centering
  \caption{Distinct graph augmentation pairs.}
    \vspace{-3pt}
     \begin{small}
\setlength{\tabcolsep}{2.5pt}
  {
  \begin{tabular}{lcccc}
   \toprule
     &$A_{i,j}$ &$A_i,A_j$ &Augmentation &Augmentation\\
     \hline
     &$A_{1,1}$ &$A_1$,$A_1$
     &Identical &Identical\\
     &$A_{1,2}$&$A_1$, $A_2$ &Identical &Node Dropping\\
     &$A_{1,3}$&$A_1$, $A_3$ &Identical &Edge Perturbation\\
     &$A_{1,4}$&$A_1$, $A_4$ &Identical &Subgraph\\
     &$A_{1,5}$&$A_1$, $A_5$ &Identical &Attribute Masking\\
     &$A_{2,2}$&$A_2$, $A_2$ &Node Dropping &Node Dropping\\
     &$A_{2,3}$&$A_2$, $A_3$ &Node Dropping &Edge Perturbation\\
     &$A_{2,4}$&$A_2$, $A_4$ &Node Dropping &Subgraph\\
     &$A_{2,5}$&$A_2$, $A_5$ &Node Dropping &Attribute Masking\\
     &$A_{3,3}$&$A_3$, $A_3$ &Edge Perturbation &Edge Perturbation\\
     &$A_{3,4}$&$A_3$, $A_4$ &Edge Perturbation &Subgraph\\
     &$A_{3,5}$&$A_3$, $A_5$ &Edge Perturbation &Attribute Masking\\
     &$A_{4,4}$&$A_4$, $A_4$ &Subgraph &Subgraph\\
     &$A_{4,5}$&$A_4$, $A_5$ &Subgraph &Attribute Masking\\
     &$A_{5,5}$&$A_5$, $A_5$ &Attribute Masking &Attribute Masking\\
 \bottomrule
\end{tabular}}
\end{small}
\label{aug_pairs}
\vspace{-6pt}
\end{table}

\section{Experiments}
\label{experiment}
We evaluate the performance of GPA on multiple graph datasets with various scales and types, and focus on exploring the following research questions.

\begin{itemize}
    \item \textbf{Q1}: How effective is GPA in performing graph representation learning against state-of-the-art GCL methods in unsupervised and semi-supervised evaluation tasks?
    \item \textbf{Q2}: How effective is the proposed personalized augmentation selector in identifying augmentations across various datasets?
    \item \textbf{Q3}: Compared with random selection, how effective is our proposed personalized augmentation selector? 
    \item \textbf{Q4}: What are the impacts of hyperparameters on GPA, such as the embedding dimension $d$ of the score function?
\end{itemize}

We first briefly introduce the experimental setup, then the details of the experimental results and their analysis.

\subsection{Datasets and Experiment Settings}
\subsubsection{Datasets}
For a comprehensive comparison, we evaluate the performance of GPA on eleven widely used benchmark datasets. Specifically, we include two small molecules networks (\textbf{NCI1} and \textbf{MUTAG}), two bioinformatics networks (\textbf{DD} and \textbf{PROTEINS}), and five social networks (\textbf{COLLAB}, \textbf{REDDIT-BINARY}, \textbf{REDDIT-MULTI-5K}, \textbf{IMDB-BINARY}, and \textbf{GITHUB}) from TUDatasets~\citep{Morris+2020}. To evaluate the scalability of our model, we also use one challenging OGB~\citep{hu2020open} dataset -- \textbf{ogbg-molhiv}. The data statistics are summarized in Table~\ref{dataset_stat}.

\begin{table}[t]
\centering
  \caption{Statistics of the datasets.}
    \vspace{-0pt}
    \begin{small}
\setlength{\tabcolsep}{2.5pt}
  {
  \begin{tabular}{c| c |c |c | c}
   \toprule
     &$\mid\mathcal{G}\mid$ &Avg.Nodes &Avg.Edges &$\# Label$\\
     \hline
     NCI1 &$4,110$ &$29.87$ &$32.30$ &$2$\\
     PROTEINS &$1,113$ &$39.06$ &$72.82$ &$2$\\
     DD &$1,178$ &$284.32$ &$715.66$ &$2$\\
     MUTAG &$188$ &$17.93$ &$19.79$ &$2$\\
     COLLAB &$5,000$ &$74.49$ &$2,457.78$ &$3$\\
     IMDB-BINARY &$1,000$ &$19.77$ &$96.53$ &$2$\\
     REDDIT-BINARY &$2,000$ &$429.63$ &$497.75$ &$2$\\
     REDDIT-MULTI-5K &$4,999$ &$508.52$ &$594.87$ &$5$\\
     GITHUB &$12,725$ &$113.79$ &$234.64$ &$2$\\
     ogbg-molhiv &$41,127$ &$25.5$ &$27.5$ &$2$ \\
 \bottomrule
\end{tabular}}
\end{small}
\label{dataset_stat}
\vspace{-0pt}
\end{table}

\subsubsection{Learning protocols}
Following common protocols~\citep{sun2019infograph,you2020graph}, we aim to evaluate the performance of GPA in unsupervised and semi-supervised settings. In our model training phase, we randomly split 10\% of graphs in each dataset into the validation set and use the remaining for training. After the model is trained, in unsupervised setting, we train an SVM classifier on the graph representations generated by the trained GCL model, and apply 10-fold cross-validation to evaluate the performance. For semi-supervised setting, we finetune the GCL model (its GNN encoder) with a logistic regression layer for semi-supervised learning, where the labeled sample ratio is 0.1. To avoid randomness, we repeat the process for ten times and report the averaged results.

\subsubsection{Implementation details}
Our model is built upon Pytorch and PyG (PyTorch Geometric) library~\citep{fey2019fast}. We train our model with Adam optimizer using a fixed batch size of 128. Similar to GraphCL~\citep{you2020graph}, the default augmentation ratio is set to 0.2 for all augmentation types. In unsupervised setting, we adopt a three-layer GIN~\citep{xu2018powerful} encoder with hidden dimension 128 for all datasets. In semi-supervised scenarios, we employ a five-layer ResGCN~\citep{chen2019powerful} encoder with dimension 128 for TUDatasets~\citep{Morris+2020}, while a five-layer GIN~\citep{xu2018powerful} encoder with dimension 300 for OGB datasets as suggested in~\citep{hu2020open}. There is one hyper-parameter in our model, i.e., the hidden dimension $d$ of score function $g_\theta$.  We search $d$ within the set \{128, 256, 512\}. The impact of hidden dimension is analyzed in Sec.~\ref{parameter_analysis}.

\subsubsection{Baseline methods}
To validate the effectiveness of GPA, we compare against three categories of state-of-the-art competitors. First, to evaluate the effectiveness of contrastive learning, we include one traditional network embedding method \textbf{GAE}~\citep{kipf2016variational}. Second, to study why we need personalized augmentation, we include classic GCL methods that assign identical augmentation strategies for all graphs
\textbf{InfoGraph}~\citep{sun2019infograph},  \textbf{GraphCL}~\citep{you2020graph}, \textbf{JOAO}~\citep{you2021graph} and its variant \textbf{JOAOv2}~\citep{you2021graph}. Third, to have a comprehensive evaluation, we also include a learning-based augmentation method \textbf{AD-GCL}~\citep{suresh2021adversarial}. AD-GCL aims to learn different adjacency matrices for different graphs, but it is limited to edge-dropping based augmentation. It cannot be applied to other advanced augmentation techniques, such as attribute mask and node dropping.

\begin{table*}[htpb]
\caption{Unsupervised learning performance for graph classification in TUdatasets (Averaged accuracy $\pm$ std. over 10 runs). The \textbf{bold} numbers denote the best performance and
the numbers in \textcolor{blue}{blue} represent the second best performance}
\vspace{-5pt}
\label{unsuper_result_table}
\begin{center}
\begin{small}
\resizebox{\linewidth}{!}{
\begin{tabular}{lcccccccccc}
\toprule
Dataset &NCI1 &PROTEINS & DD & MUTAG &COLLAB &RDT-B &RDT-M5K&IMDB-B\\
\midrule

InfoGraph &$76.20\pm 1.06$
&$74.44\pm 0.31$
&$72.85\pm 1.78$
&\textcolor{blue}{$89.01\pm 1.13$} 
&$70.65\pm 1.13$
&$82.50\pm 1.42$
&$53.46\pm 1.03$ 
&\textcolor{blue}{$73.03\pm 0.87$}\\
GraphCL &$77.87\pm 0.41$ 
&$74.39\pm 0.45$ 
&\textcolor{blue}{$78.62\pm 0.40$} 
&$86.80\pm 1.34$ 
&$71.36\pm 1.15$ 
&\bm{$89.53\pm 0.84$}
&\textcolor{blue}{$55.99\pm 0.28$} 
&$71.14\pm 0.44$\\

JOAO &$78.07\pm 0.47$ 
&\textcolor{blue}{$74.55\pm 0.41$} 
&$77.32\pm 0.54$ 
&$87.35\pm 1.02$ 
&$69.50\pm 0.36$ 
&$85.29\pm 1.35$ 
&$55.74\pm 0.63$ 
&$70.21\pm 3.08$\\
JOAOv2 &\textcolor{blue}{$78.36\pm 0.53$ }
&$74.07\pm 1.10$ 
&$77.40\pm 1.15$ 
&$87.67\pm 0.79$ 
&$69.33\pm 0.34$ 
&$86.42\pm 1.45$ 
&\bm{$56.03\pm 0.27$} 
&$70.83\pm 0.25$\\
\midrule
AD-GCL &$69.67\pm 0.51$
&$73.59\pm 0.65$
&$74.49\pm 0.52$
&$88.62\pm 1.27$
&\textcolor{blue}{$73.32\pm 0.61$}
&$85.52\pm 0.79$
&$53.00\pm 0.82$
&$71.57\pm 1.01$
\\
\midrule

GPA &\bm{$80.42\pm 0.41$}
&\bm{$75.94\pm 0.25$}
&\bm{$79.90\pm 0.35$}
&\bm{$89.68\pm 0.80$}
&\bm{$76.17\pm 0.10$}
&\textcolor{blue}{$89.32\pm 0.38$}
&$53.70\pm 0.19$
&\bm{$74.64\pm 0.35$}

\\

\bottomrule
\end{tabular}
}
\end{small}
\vspace{-8pt}
\end{center}
\end{table*}

\begin{table*}[htpb]
\caption{Semi-supervised learning performance for graph classification in TUdatasets.}
\vspace{-5pt}
\label{semi_result_table}
\begin{center}
\begin{small}
\resizebox{\linewidth}{!}{
\begin{tabular}{lcccccccccc}
\toprule
Dataset &NCI1 &PROTEINS & DD &COLLAB &RDT-B &RDT-M5K&GITHUB\\
\midrule
GAE &$74.36\pm 0.24$
&$70.51\pm 0.17$
&$74.54\pm 0.68$
&$75.09\pm 0.19$
&$87.69\pm 0.40$
&$53.58\pm 0.13$
&$63.89\pm 0.52$\\
\midrule
InfoGraph &$74.86\pm 0.26$
&$72.27\pm 0.40$
&$75.78\pm 0.34$
&$73.76\pm 0.29$
&$88.66\pm 0.95$
&\textcolor{blue}{$53.61\pm 0.31$}
&$65.21\pm 0.88$
\\

GraphCL &$74.63\pm 0.25$
&\textcolor{blue}{$74.17\pm 0.34$}
&$76.17\pm 1.37$
&$74.23\pm 0.21$
&\textcolor{blue}{$89.11\pm 0.19$}
&$52.55\pm 0.45$
&$65.81\pm 0.79$\\

JOAO &$74.48\pm 0.25$
&$72.13\pm 0.92$
&$75.69\pm 0.67$
&$75.30\pm 0.32$
&$88.14\pm 0.25$
&$52.83\pm 0.54$
&$65.00\pm 0.30$\\
JOAOv2 &$74.86\pm 0.39$
&$73.31\pm 0.48$
&$75.81\pm 0.73$
&\textcolor{blue}{$75.53\pm 0.18$}
&$88.79\pm 0.65$
&$52.71\pm 0.28$
&$66.60\pm 0.60$\\
\midrule
AD-GCL &\textcolor{blue}{$75.18\pm 0.31$}
&$73.96\pm 0.47$
&\bm{$77.91\pm 0.73$}
&\bm{$75.82\pm 0.26$}
&\bm{$90.10\pm 0.15$}
&$53.49\pm 0.28$
&\textcolor{blue}{$67.13\pm 0.52$}
\\

\midrule

GPA &\bm{$75.50\pm 0.14$}
&\bm{$74.27\pm 1.11$}
&\textcolor{blue}{$76.68\pm 0.81$}
&$70.45\pm 0.06$
&\bm{$89.99\pm 0.32$}
&\bm{$54.92\pm 0.35$}
&\bm{$68.31\pm 0.13$}
\\

\bottomrule
\end{tabular}
}
\end{small}
\vspace{-8pt}
\end{center}
\end{table*}

\subsection{Comparison with Baselines}
\subsubsection{On diverse datasets from TUDataset}
We start by comparing the performance of GPA with the state-of-the-art baseline methods under two settings (\textbf{Q1}). Table~\ref{unsuper_result_table} \& Table~\ref{semi_result_table} report the results of all methods on unsupervised and semi-supervised settings, respectively. Through comparison between (1) GPA and vanilla GCL methods and (2) GPA and augmentation learnable based GCL method, we have the following \textbf{Obs}ervations.

\textbf{Obs.1. With personalized augmentations for each graph, GPA outperforms vanilla GCL methods with fixed augmentation per dataset.} By identifying different augmentations for different sample graphs, GPA performs generally better than vanilla GCL methods on two evaluation scenarios (Table~\ref{unsuper_result_table} and Table~\ref{semi_result_table}). Specifically, in the semi-supervised evaluation task, GPA consistently outperforms GAE, InfoGraph, GraphCL, JOAO, and JOAOv2 on all datasets. In the unsupervised setting, GPA outperforms the vanilla GCL methods on 6 out of 8 datasets. 
In particular, GPA improves 7.8\%,  9.9\%, 9.6\%, and 6.7\% over InfoGraph, JOAOv2, JOAO, and GraphCL on COLLAB in Table~\ref{unsuper_result_table}, respectively. This observation validates the effectiveness of performing personalized augmentation in GCL training.

\begin{wraptable}{r}{4.3cm}
    \centering
    \caption{Semi-supervised learning performance on large-scale OGB datasets.}
  \vspace{-3pt}
\begin{center}
\begin{small}
\setlength{\tabcolsep}{2.45pt}
\begin{tabular}{lccc}
\toprule
 & 
 &ogbg-molhiv\\
\midrule
&GraphCL &$55.48\pm 1.32$\\
&JOAOv2 &$57.39\pm 1.39$\\
&GPA &$60.76\pm 1.01$
\\
\bottomrule
\end{tabular}
\end{small}
\end{center}
\vspace{-10pt}
\label{results_ogb}
\end{wraptable}



\textbf{Obs.2. Across diverse datasets, GPA performs better than
(or on par with) the learnable GCL methods on two evaluation settings.} On all datasets originating from diverse domains, GPA generally performs better or sometimes on par with the state-of-the-art AD-GCL method. In unsupervised setting (Table~\ref{unsuper_result_table}), GPA consistently outperforms AD-GCL on all datasets. Specifically, GPA could improve up to 15.4\% over AD-GCL on NCI and DD datasets. In semi-supervised setting, GPA beats or matches with AD-GCL on 5 out of 7 datasets. 
The main difference between AD-GCL and our model is that the former adopts the edge dropping augmentation type for all graphs but under different tailored dropping details. In contrast, our model focuses on identifying the optimal augmentation types for each graph from a set of different augmentation types. Thus, for those datasets (i.e., DD and COLLAB) that our model loses to AD-GCL, they may prefer edge dropping augmentation type more than other graph augmentation strategies, thereby naturally achieving better performance than ours since they have tailored augmented details under edge dropping augmentation. These results demonstrate our motivation to assign various augmentation types for different graph instances, since various graphs may favor various augmentation types, e.g., attribute masking.

\subsubsection{On large-scale OGB datasets} To study the scalability of our model, we further conduct experiments on one large-scale OGB datasets. Table~\ref{results_ogb} reports the results of our model and the most related baseline GraphCL (AD-GCL is excluded for its poor scalability). We have the following observations.

\textbf{Obs.3. GPA scales well on large datasets.} 
GPA consistently performs better than GraphCL and JOAOv2 (See Table~\ref{results_ogb}). 
To be specific, GPA improves 9.5\% and 5.9\% over GraphCL and JOAOv2 on ogbg-molhiv, respectively.

\begin{figure}[h]
  \centering
  \includegraphics[width=14cm]{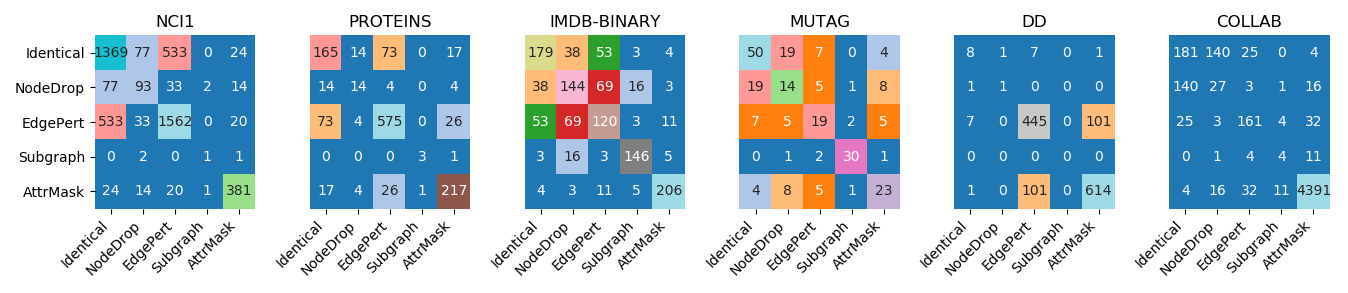}
  \vspace{-6pt}
  \caption{Augmentation distribution learned by GPA over molecules, bioinformatics, and social networks, in terms of the unsupervised setting.
  }
  \label{figure_aug}
\end{figure}


\subsection{Personalized Augmentation Analysis}
To study the effectiveness of our model in identifying informative augmentation types for various graphs (\textbf{Q2}), we visualize the learned augmentation distribution on Fig.~\ref{figure_aug}. By comparing across different types of datasets, we observe the following.

\textbf{Obs.5. By learning from the data, GPA can effectively assign different augmentations for various datasets.} Our model GPA can identify different augmentations for different sample graphs, and allow different datasets to have their own augmentation distributions (see Fig.~\ref{figure_aug}). Specifically, on MUTAG, 19 graphs prefer (\textit{Identical, NodeDrop}) augmentations, while 30 graphs favor (\textit{Subgraph, Subgraph}) augmentation combinations. Notice that (\textit{Subgraph, Subgraph}) will generate two different subgraph-perturbation-induced augmented views, owning to sample randomness. Besides, COLLAB more likes the (\textit{AttMask, AttMask}) augmentation pair, while DD prefers (\textit{EdgePert, EdgePert}) operations. These observations empirically echo the necessity of performing personalized augmentation for GCL methods.

Another promising observation is that our model can assign (\textit{Identical, Identical}) choice (i.e., two identical views) to some portion of graphs over all datasets. Given that the mutual information between two identical views (i.e., representations) is always maximized, such pure identical augmentations can be regarded as a skip operation. That is, these graphs abandon themselves during the GCL model training. The possible reason is that the existing augmentation strategies are not suitable to capture their characteristics or damage their semantic meanings. In this case, blindly selecting any combination of other augmentation types may incur huge performance degradation or noise. This observation sheds light on designing more advanced augmentation strategies beyond the current basic augmentations. On the other hand, it verifies the effectiveness of the proposed augmentation selector in skipping noisy graphs during model training by providing identical augmentations.

\begin{table}[t]
\caption{Ablation study of GPA under unsupervised setting in terms of mean classification accuracy results (in \%).}
\label{table_ablation}
  \vspace{-3pt}
\begin{center}
\begin{small}
\setlength{\tabcolsep}{2.45pt}
\begin{tabular}{lccccc}
\toprule
 & 
 &GPA-random 
 &GPA\\
\midrule
&NCI1 
&$77.71\pm 0.60$
&\bm{$80.42\pm 0.41$}\\
&PROTEINS 
&$74.21\pm 0.39$
&\bm{$75.94\pm 0.25$}\\
&DD 
&$77.25\pm 0.69$
&\bm{$79.90\pm 0.35$}\\
&MUTAG 
&$86.08\pm 2.93$
&\bm{$89.68\pm 0.80$}\\
&IMDB-B 
&$71.80\pm 1.13$
&\bm{$74.64\pm 0.35$}\\
\bottomrule
\end{tabular}
\end{small}
\end{center}
\vspace{-10pt}
\end{table}

\subsection{Ablation Study}
\label{ablation}

To further investigate the effectiveness of the proposed personalized augmentation selector (\textbf{Q3}), we compare it with a random-search based variant, i.e., GPA-random. GPA-random replaces the personalized augmentation selector with a random mechanism. Specifically, it assigns one random augmentation pair to each graph. Noticed that GPA-random still assigns different augmentations to each graph while GraphCL assigns one pre-defined pair of augmentation strategies to the whole dataset.
Table~\ref{table_ablation} shows the results in terms of unsupervised setting. From the table, we can observe that GPA consistently performs better than GPA-random in all cases. In particular, GPA improves 3.5\%, 2.3\%, 3.4\%, 4.2\%, and 4.0\% over GPA-random on NCI1, PROTEINS, DD, MUTAG, and IMDB-B, respectively. This comparison validates our motivation to develop a tailored and learnable personalized augmentation selector for GCL methods.

\begin{wrapfigure}{r}{4.7cm}
\vspace{-50pt}
\includegraphics[width=4.7cm]{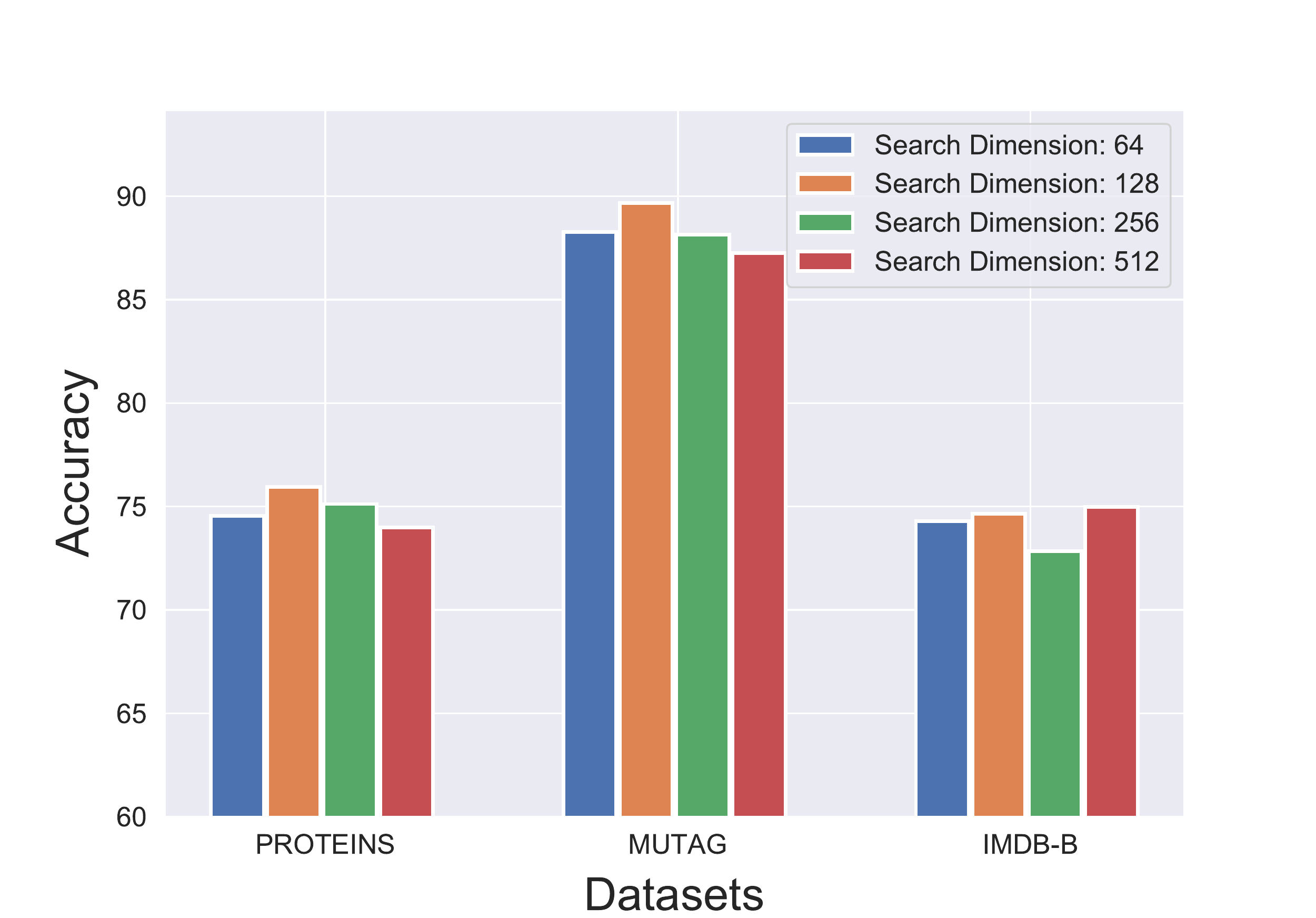}
  \caption{Personalized augmentation selector dimension analysis of GPA} 
  \label{figure_searchdim}
  \vspace{-50pt}
\end{wrapfigure}

\subsection{Parameter Sensitivity Analysis}
\label{parameter_analysis}

We now study the impact of parameter, i.e., the hidden dimension $d$ of the score function, to answer (\textbf{Q4}). Specifically, we search $d$ from the set \{64, 128, 256, 512\} and plot the results of GPA on three representative datasets in Fig.~\ref{figure_searchdim}. Similar observations are obtained by other datasets. From the figure, we can see that GPA generally performs stably across various dimension choices. In experiments, we fix $d=128$ for all datasets.

\section{Conclusions and Future Work}
In this paper, we study the augmentation selection problem for graph contrastive learning. Existing GCL efforts mainly focus on employing two shared augmentation strategies for all graphs in the dataset based on the assumption that they contain similar nature. Here, we argue that such collective augmentation selection is suboptimal in practice due to the heterogeneity of graph structure data. Different graphs should have different augmentation preferences. To bridge the gap, we propose a novel graph contrastive learning framework with personalized augmentation termed as GPA. GPA not only allows each graph to select its optimal augmentation types, but also automates the selection via a personalized augmentation selector, which can be jointly trained with downstream GCL models under a bi-level optimization. Empirical results on the graph classification task demonstrate the superiority of GPA against state-of-the-art GCL methods in terms of unsupervised and semi-supervised settings, across multiple benchmark datasets with various types such as molecules, bioinformatics, and social networks. In the future, we will explore how to extend our personalized augmentation idea to the node-level contrastive learning task. Moreover, it would be interesting to investigate whether our model works for other data types, such as image or text data.

\bibliography{iclr2023_conference}

\begin{thebibliography}{46}
\providecommand{\natexlab}[1]{#1}
\providecommand{\url}[1]{\texttt{#1}}
\expandafter\ifx\csname urlstyle\endcsname\relax
  \providecommand{\doi}[1]{doi: #1}\else
  \providecommand{\doi}{doi: \begingroup \urlstyle{rm}\Url}\fi

\bibitem[Chen et~al.(2022)Chen, Lin, Li, Li, Zhou, Sun,
  et~al.]{chen2020distance}
Deli Chen, Yanyai Lin, Lei Li, Xuancheng~Ren Li, Jie Zhou, Xu~Sun, et~al.
\newblock Distance-wise graph contrastive learning.
\newblock \emph{IJCAI}, 2022.

\bibitem[Chen et~al.(2018)Chen, Ma, and Xiao]{chen2018fastgcn}
Jie Chen, Tengfei Ma, and Cao Xiao.
\newblock Fastgcn: Fast learning with graph convolutional networks via
  importance sampling.
\newblock In \emph{ICLR}, 2018.

\bibitem[Ding et~al.(2018)Ding, Tang, and Zhang]{ding2018semi}
Ming Ding, Jie Tang, and Jie Zhang.
\newblock Semi-supervised learning on graphs with generative adversarial nets.
\newblock In \emph{CIKM}, pp.\  913--922, 2018.

\bibitem[Fey \& Lenssen(2019)Fey and Lenssen]{fey2019fast}
Matthias Fey and Jan~E. Lenssen.
\newblock Fast graph representation learning with {PyTorch Geometric}.
\newblock In \emph{ICLR Workshop on Representation Learning on Graphs and
  Manifolds}, 2019.

\bibitem[Garcia~Duran \& Niepert(2017)Garcia~Duran and
  Niepert]{garcia2017learning}
Alberto Garcia~Duran and Mathias Niepert.
\newblock Learning graph representations with embedding propagation.
\newblock \emph{NeurIPS}, 30, 2017.

\bibitem[Gilmer et~al.(2017)Gilmer, Schoenholz, Riley, Vinyals, and
  Dahl]{gilmer2017neural}
Justin Gilmer, Samuel~S Schoenholz, Patrick~F Riley, Oriol Vinyals, and
  George~E Dahl.
\newblock Neural message passing for quantum chemistry.
\newblock In \emph{ICML}, pp.\  1263--1272. PMLR, 2017.

\bibitem[Hamilton et~al.(2017)Hamilton, Ying, and
  Leskovec]{hamilton2017inductive}
Will Hamilton, Zhitao Ying, and Jure Leskovec.
\newblock Inductive representation learning on large graphs.
\newblock \emph{NeurIPS}, 30, 2017.

\bibitem[Hassani \& Khasahmadi(2020)Hassani and
  Khasahmadi]{hassani2020contrastive}
Kaveh Hassani and Amir~Hosein Khasahmadi.
\newblock Contrastive multi-view representation learning on graphs.
\newblock In \emph{ICML}, pp.\  4116--4126. PMLR, 2020.

\bibitem[Hou et~al.(2022)Hou, Liu, Dong, Wang, Tang, et~al.]{hou2022graphmae}
Zhenyu Hou, Xiao Liu, Yuxiao Dong, Chunjie Wang, Jie Tang, et~al.
\newblock Graphmae: Self-supervised masked graph autoencoders.
\newblock \emph{arXiv preprint arXiv:2205.10803}, 2022.

\bibitem[Hu et~al.(2020{\natexlab{a}})Hu, Fey, Zitnik, Dong, Ren, Liu, Catasta,
  and Leskovec]{hu2020open}
Weihua Hu, Matthias Fey, Marinka Zitnik, Yuxiao Dong, Hongyu Ren, Bowen Liu,
  Michele Catasta, and Jure Leskovec.
\newblock Open graph benchmark: Datasets for machine learning on graphs.
\newblock \emph{NeurIPS}, 33:\penalty0 22118--22133, 2020{\natexlab{a}}.

\bibitem[Hu et~al.(2020{\natexlab{b}})Hu, Dong, Wang, Chang, and
  Sun]{hu2020gpt}
Ziniu Hu, Yuxiao Dong, Kuansan Wang, Kai-Wei Chang, and Yizhou Sun.
\newblock Gpt-gnn: Generative pre-training of graph neural networks.
\newblock In \emph{KDD}, pp.\  1857--1867, 2020{\natexlab{b}}.

\bibitem[Iyer et~al.(2021)Iyer, Wang, and Sun]{iyer2021bi}
Roshni~G Iyer, Wei Wang, and Yizhou Sun.
\newblock Bi-level attention graph neural networks.
\newblock In \emph{ICDM}, pp.\  1126--1131. IEEE, 2021.

\bibitem[Jiao et~al.(2020)Jiao, Xiong, Zhang, Zhang, Zhang, and
  Zhu]{jiao2020sub}
Yizhu Jiao, Yun Xiong, Jiawei Zhang, Yao Zhang, Tianqi Zhang, and Yangyong Zhu.
\newblock Sub-graph contrast for scalable self-supervised graph representation
  learning.
\newblock In \emph{ICDM}, pp.\  222--231. IEEE, 2020.

\bibitem[Jin et~al.(2020)Jin, Derr, Liu, Wang, Wang, Liu, and
  Tang]{jin2020self}
Wei Jin, Tyler Derr, Haochen Liu, Yiqi Wang, Suhang Wang, Zitao Liu, and
  Jiliang Tang.
\newblock Self-supervised learning on graphs: Deep insights and new direction.
\newblock \emph{arXiv preprint arXiv:2006.10141}, 2020.

\bibitem[Kipf \& Welling(2016)Kipf and Welling]{kipf2016variational}
Thomas~N Kipf and Max Welling.
\newblock Variational graph auto-encoders.
\newblock \emph{NeurIPS}, 2016.

\bibitem[Kipf \& Welling(2017)Kipf and Welling]{kipf2017semi}
Thomas~N. Kipf and Max Welling.
\newblock Semi-supervised classification with graph convolutional networks.
\newblock In \emph{ICLR}, 2017.

\bibitem[Liu et~al.(2018)Liu, Simonyan, and Yang]{liu2018darts}
Hanxiao Liu, Karen Simonyan, and Yiming Yang.
\newblock Darts: Differentiable architecture search.
\newblock In \emph{ICLR}, 2018.

\bibitem[Liu et~al.(2019)Liu, Tan, Li, Yang, Zhou, and Hu]{liu2019single}
Ninghao Liu, Qiaoyu Tan, Yuening Li, Hongxia Yang, Jingren Zhou, and Xia Hu.
\newblock Is a single vector enough? exploring node polysemy for network
  embedding.
\newblock In \emph{Proceedings of the 25th ACM SIGKDD International Conference
  on Knowledge Discovery \& Data Mining}, pp.\  932--940, 2019.

\bibitem[Liu et~al.(2021)Liu, Zhang, Hou, Mian, Wang, Zhang, and
  Tang]{liu2021self}
Xiao Liu, Fanjin Zhang, Zhenyu Hou, Li~Mian, Zhaoyu Wang, Jing Zhang, and Jie
  Tang.
\newblock Self-supervised learning: Generative or contrastive.
\newblock \emph{TKDE}, 2021.

\bibitem[Morris et~al.(2020)Morris, Kriege, Bause, Kersting, Mutzel, and
  Neumann]{Morris+2020}
Christopher Morris, Nils~M. Kriege, Franka Bause, Kristian Kersting, Petra
  Mutzel, and Marion Neumann.
\newblock Tudataset: A collection of benchmark datasets for learning with
  graphs.
\newblock In \emph{ICML 2020 Workshop on Graph Representation Learning and
  Beyond (GRL+ 2020)}, 2020.
\newblock URL \url{www.graphlearning.io}.

\bibitem[Peng et~al.(2020)Peng, Huang, Luo, Zheng, Rong, Xu, and
  Huang]{peng2020graph}
Zhen Peng, Wenbing Huang, Minnan Luo, Qinghua Zheng, Yu~Rong, Tingyang Xu, and
  Junzhou Huang.
\newblock Graph representation learning via graphical mutual information
  maximization.
\newblock In \emph{WWW}, pp.\  259--270, 2020.

\bibitem[Qiu et~al.(2020)Qiu, Chen, Dong, Zhang, Yang, Ding, Wang, and
  Tang]{qiu2020gcc}
Jiezhong Qiu, Qibin Chen, Yuxiao Dong, Jing Zhang, Hongxia Yang, Ming Ding,
  Kuansan Wang, and Jie Tang.
\newblock Gcc: Graph contrastive coding for graph neural network pre-training.
\newblock In \emph{SIGKDD}, pp.\  1150--1160, 2020.

\bibitem[Shaban et~al.(2019)Shaban, Cheng, Hatch, and
  Boots]{shaban2019truncated}
Amirreza Shaban, Ching-An Cheng, Nathan Hatch, and Byron Boots.
\newblock Truncated back-propagation for bilevel optimization.
\newblock In \emph{AISTATS}, pp.\  1723--1732. PMLR, 2019.

\bibitem[Snoek et~al.(2012)Snoek, Larochelle, and Adams]{snoek2012practical}
Jasper Snoek, Hugo Larochelle, and Ryan~P Adams.
\newblock Practical bayesian optimization of machine learning algorithms.
\newblock \emph{NeurIPS}, 25, 2012.

\bibitem[Sun et~al.(2020)Sun, Hoffmann, Verma, and Tang]{sun2019infograph}
Fan-Yun Sun, Jordan Hoffmann, Vikas Verma, and Jian Tang.
\newblock Infograph: Unsupervised and semi-supervised graph-level
  representation learning via mutual information maximization.
\newblock \emph{ICLR}, 2020.

\bibitem[Suresh et~al.(2021)Suresh, Li, Hao, and
  Neville]{suresh2021adversarial}
Susheel Suresh, Pan Li, Cong Hao, and Jennifer Neville.
\newblock Adversarial graph augmentation to improve graph contrastive learning.
\newblock \emph{NeurIPS}, 34, 2021.

\bibitem[Tan et~al.(2019)Tan, Liu, and Hu]{tan2019deep}
Qiaoyu Tan, Ninghao Liu, and Xia Hu.
\newblock Deep representation learning for social network analysis.
\newblock \emph{Frontiers in big Data}, 2:\penalty0 2, 2019.

\bibitem[Tan et~al.(2020)Tan, Liu, Zhao, Yang, Zhou, and Hu]{tan2020learning}
Qiaoyu Tan, Ninghao Liu, Xing Zhao, Hongxia Yang, Jingren Zhou, and Xia Hu.
\newblock Learning to hash with graph neural networks for recommender systems.
\newblock In \emph{Proceedings of The Web Conference 2020}, pp.\  1988--1998,
  2020.

\bibitem[Tan et~al.(2022)Tan, Liu, Huang, Chen, Choi, and Hu]{tan2022mgae}
Qiaoyu Tan, Ninghao Liu, Xiao Huang, Rui Chen, Soo-Hyun Choi, and Xia Hu.
\newblock Mgae: Masked autoencoders for self-supervised learning on graphs.
\newblock \emph{arXiv preprint arXiv:2201.02534}, 2022.

\bibitem[Thakoor et~al.(2021)Thakoor, Tallec, Azar, Munos,
  Veli{\v{c}}kovi{\'c}, and Valko]{thakoor2021bootstrapped}
Shantanu Thakoor, Corentin Tallec, Mohammad~Gheshlaghi Azar, R{\'e}mi Munos,
  Petar Veli{\v{c}}kovi{\'c}, and Michal Valko.
\newblock Bootstrapped representation learning on graphs.
\newblock In \emph{ICLR 2021 Workshop on Geometrical and Topological
  Representation Learning}, 2021.

\bibitem[Ting~Chen(2019)]{chen2019powerful}
Yizhou~Sun Ting~Chen, Song~Bian.
\newblock Are powerful graph neural nets necessary? a dissection on graph
  classification.
\newblock \emph{CoRR}, abs/1905.04579, 2019.

\bibitem[Veli{\v{c}}kovi{\'c} et~al.(2018)Veli{\v{c}}kovi{\'c}, Cucurull,
  Casanova, Romero, Li{\`o}, and Bengio]{velivckovic2017graph}
Petar Veli{\v{c}}kovi{\'c}, Guillem Cucurull, Arantxa Casanova, Adriana Romero,
  Pietro Li{\`o}, and Yoshua Bengio.
\newblock Graph attention networks.
\newblock In \emph{ICLR}, 2018.

\bibitem[Velickovic et~al.(2019)Velickovic, Fedus, Hamilton, Li{\`o}, Bengio,
  and Hjelm]{velickovic2019deep}
Petar Velickovic, William Fedus, William~L Hamilton, Pietro Li{\`o}, Yoshua
  Bengio, and R~Devon Hjelm.
\newblock Deep graph infomax.
\newblock \emph{ICLR (Poster)}, 2\penalty0 (3):\penalty0 4, 2019.

\bibitem[Veli{\v{c}}kovi{\'c} et~al.(2019)Veli{\v{c}}kovi{\'c}, Fedus,
  Hamilton, Li{\`o}, Bengio, and Hjelm]{velivckovic2018deep}
Petar Veli{\v{c}}kovi{\'c}, William Fedus, William~L Hamilton, Pietro Li{\`o},
  Yoshua Bengio, and R~Devon Hjelm.
\newblock Deep graph infomax.
\newblock \emph{ICLR}, 2019.

\bibitem[Wan et~al.(2020)Wan, Pan, Yang, and Gong]{wan2020contrastive}
Sheng Wan, Shirui Pan, Jian Yang, and Chen Gong.
\newblock Contrastive and generative graph convolutional networks for
  graph-based semi-supervised learning.
\newblock \emph{AAAI}, 2020.

\bibitem[Wang et~al.(2022)Wang, Zhang, Zhu, and Huang]{wang2022augmentation}
Haonan Wang, Jieyu Zhang, Qi~Zhu, and Wei Huang.
\newblock Augmentation-free graph contrastive learning.
\newblock \emph{arXiv preprint arXiv:2204.04874}, 2022.

\bibitem[Wang et~al.(2019{\natexlab{a}})Wang, Zhang, Liu, Chen, Xu, Fardad, and
  Li]{wang2019towards}
Jingkang Wang, Tianyun Zhang, Sijia Liu, Pin-Yu Chen, Jiacen Xu, Makan Fardad,
  and Bo~Li.
\newblock Towards a unified min-max framework for adversarial exploration and
  robustness.
\newblock \emph{arXiv preprint arXiv:1906.03563}, 2019{\natexlab{a}}.

\bibitem[Wang et~al.(2019{\natexlab{b}})Wang, Yu, Wang, Cheng, Zhang, Zha, He,
  and Chen]{wang2019learning}
Lu~Wang, Wenchao Yu, Wei Wang, Wei Cheng, Wei Zhang, Hongyuan Zha, Xiaofeng He,
  and Haifeng Chen.
\newblock Learning robust representations with graph denoising policy network.
\newblock In \emph{ICDM}, pp.\  1378--1383. IEEE, 2019{\natexlab{b}}.

\bibitem[West et~al.(2001)]{west2001introduction}
Douglas~Brent West et~al.
\newblock \emph{Introduction to graph theory}, volume~2.
\newblock Prentice hall Upper Saddle River, 2001.

\bibitem[Wu et~al.(2021)Wu, Lin, Tan, Gao, and Li]{wu2021self}
Lirong Wu, Haitao Lin, Cheng Tan, Zhangyang Gao, and Stan~Z Li.
\newblock Self-supervised learning on graphs: Contrastive, generative, or
  predictive.
\newblock \emph{IEEE TKDE}, 2021.

\bibitem[Wu et~al.(2020)Wu, Pan, Chen, Long, Zhang, and
  Philip]{wu2020comprehensive}
Zonghan Wu, Shirui Pan, Fengwen Chen, Guodong Long, Chengqi Zhang, and S~Yu
  Philip.
\newblock A comprehensive survey on graph neural networks.
\newblock \emph{TNNLS}, 32\penalty0 (1):\penalty0 4--24, 2020.

\bibitem[Xu et~al.(2018)Xu, Hu, Leskovec, and Jegelka]{xu2018powerful}
Keyulu Xu, Weihua Hu, Jure Leskovec, and Stefanie Jegelka.
\newblock How powerful are graph neural networks?
\newblock In \emph{ICLR}, 2018.

\bibitem[You et~al.(2020)You, Chen, Sui, Chen, Wang, and Shen]{you2020graph}
Yuning You, Tianlong Chen, Yongduo Sui, Ting Chen, Zhangyang Wang, and Yang
  Shen.
\newblock Graph contrastive learning with augmentations.
\newblock \emph{NeurIPS}, 33:\penalty0 5812--5823, 2020.

\bibitem[You et~al.(2021)You, Chen, Shen, and Wang]{you2021graph}
Yuning You, Tianlong Chen, Yang Shen, and Zhangyang Wang.
\newblock Graph contrastive learning automated.
\newblock \emph{ICML}, 2021.

\bibitem[Zeng \& Xie(2021)Zeng and Xie]{zeng2020contrastive}
Jiaqi Zeng and Pengtao Xie.
\newblock Contrastive self-supervised learning for graph classification.
\newblock In \emph{AAAI}, volume~35, pp.\  10824--10832, 2021.

\bibitem[Zhu et~al.(2021)Zhu, Xu, Yu, Liu, Wu, and Wang]{zhu2021graph}
Yanqiao Zhu, Yichen Xu, Feng Yu, Qiang Liu, Shu Wu, and Liang Wang.
\newblock Graph contrastive learning with adaptive augmentation.
\newblock In \emph{WWW}, pp.\  2069--2080, 2021.

\end{thebibliography}
\bibliographystyle{iclr2023_conference}


\end{document}